\documentclass[sigconf]{acmart}
\AtBeginDocument{%
  \providecommand\BibTeX{{%
    \normalfont B\kern-0.5em{\scshape i\kern-0.25em b}\kern-0.8em\TeX}}}

\renewcommand\footnotetextcopyrightpermission[1]{}
\newcommand*{\affaddr}[1]{#1} 
\newcommand*{\affmark}[1][*]{\textsuperscript{#1}}
\renewcommand*{\email}[1]{\texttt{#1}}
\newcommand\blfootnote[1]{%
  \begingroup
  \renewcommand\thefootnote{}\footnote{#1}%
  \addtocounter{footnote}{-1}%
  \endgroup
}

\usepackage{threeparttable}
\usepackage{makecell}

\usepackage{times}
\usepackage{moreverb}
\usepackage{graphicx}
\usepackage{mathrsfs}
\usepackage{bm}
\usepackage{url}
\usepackage{amsmath}
\usepackage{multirow}

\usepackage{amssymb}

\begin{document}

\title{NLUT: Neural-based 3D Lookup Tables for Video Photorealistic Style Transfer}

\author{%
Yaosen Chen\affmark[1], Han Yang\affmark[1], Yuexin Yang\affmark[1], Yuegen Liu\affmark[1], Wei Wang\affmark[1,2$\ast$], Xuming Wen\affmark[1,2], Chaoping Xie\affmark[1,2]\\
\affaddr{\affmark[1]Media Intelligence Laboratory, Sobey Digital Technology Co., Ltd} \quad \affaddr{\affmark[2]Peng Cheng Laboratory}\\
\{chenyaosen, yanhan,  yangyuexin, liuyuegen, wangwei, wenxuming, xiechaoping\}@sobey.com,\\ 
}

\renewcommand{\shortauthors}{Yaosen Chen, Han Yang, Yuexin Yang, Yuegen Liu, Wei Wang, Xuming Wen, Chaoping Xie}

\begin{abstract}
Video photorealistic style transfer is desired to generate videos with a similar photorealistic style to the style image while maintaining temporal consistency. However, existing methods obtain stylized video sequences by performing frame-by-frame photorealistic style transfer, which is inefficient and does not ensure the temporal consistency of the stylized video. To address this issue, we use neural network-based 3D Lookup Tables (LUTs) for the photorealistic transfer of videos, achieving a balance between efficiency and effectiveness. We first train a neural network for generating photorealistic stylized 3D LUTs on a large-scale dataset; then, when performing photorealistic style transfer for a specific video, we select a keyframe and style image in the video as the data source and fine-turn the neural network; finally, we query the 3D LUTs generated by the fine-tuned neural network for the colors in the video, resulting in a super-fast photorealistic style transfer, even processing $8$K video takes less than \textbf{2} millisecond per frame. The experimental results show that our method not only realizes the photorealistic style transfer of arbitrary style images but also outperforms the existing methods in terms of visual quality and consistency.  Project
page:\href{https://semchan.github.io/NLUT_Project/}{https://semchan.github.io/NLUT\_Project/}.
  
\blfootnote{$\ast$\;,Corresponding author.}

\end{abstract}

\begin{teaserfigure}
	\includegraphics[width=\textwidth]{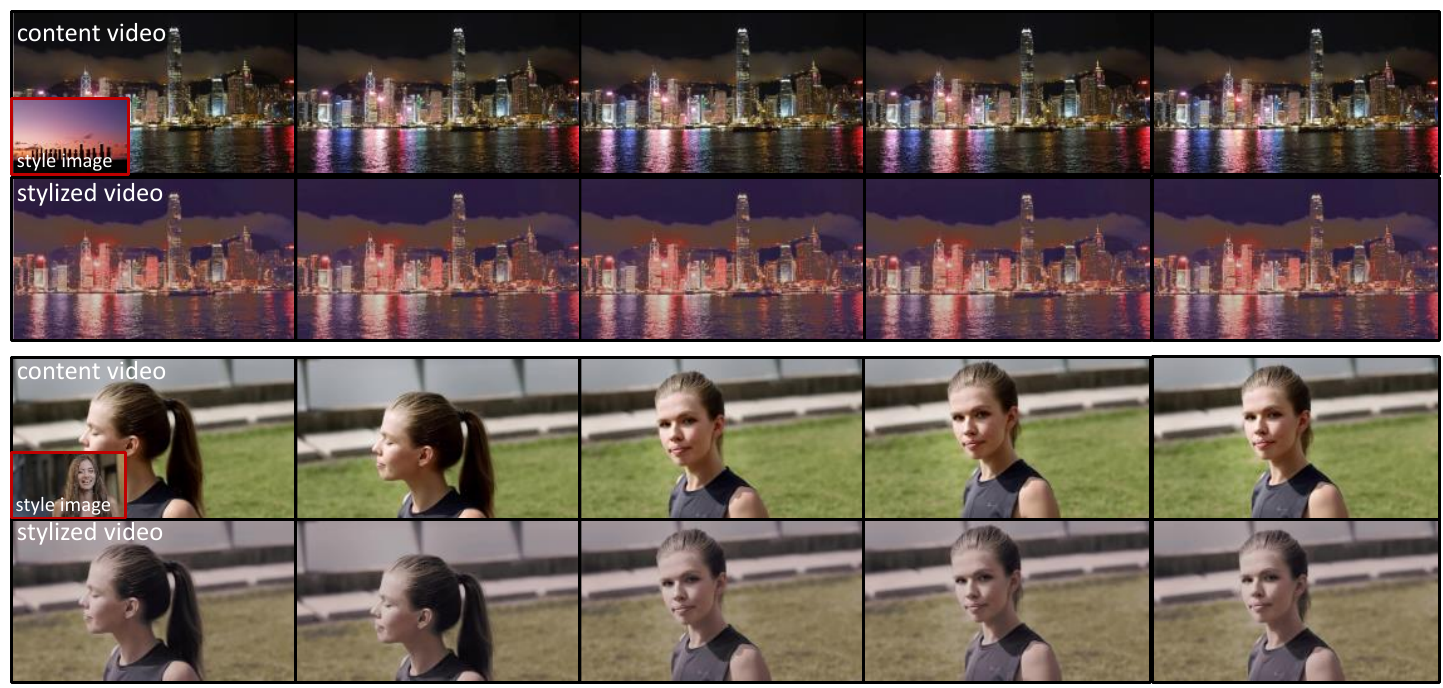}
	\caption{\textbf{Transferring photorealistic style with a style image for the video}. Given a content video and a style image, our method is able to efficiently generate photorealistic stylized video.}
	\label{fig:teaser}
\end{teaserfigure}
\maketitle

\section{Introduction}
\label{sec:intro}
The process of video photorealistic style transfer involves producing a photorealistic stylized output by rendering a video (known as the ``content video'') in the visual style of a reference image (known as the ``style image''). Fig.~\ref{fig:teaser} shows the examples. Video photorealistic style transfer has many applications, including film and television production, advertising, social media, art exhibitions, and game development. Furthermore, adding different photorealistic styles to videos can enhance the visual appeal and uniqueness of the content. Therefore, this technique has gained significant attention in recent years. 

Previous works have primarily focused on images~\cite{gatys2015neural,huang2017arbitrary,luan2017deep,li2018closed,yoo2019photorealistic,xia2020joint,qiao2021efficient,kolkin2019style,yoo2021sinir,kim2022deep,chiu2022pca} or the 3D scenes~\cite{chen2022upst,huang2022stylizednerf,chiang2022stylizing,nguyen2022snerf,zhang2022arf}, generating photorealistic images by combining content and style images. Although these methods of focusing images can generate satisfactory results for still images, when these methods are applied directly to video style transfer with frame-by-frame, they often result in color flickering due to structural changes caused by motion. Chen et al.~\cite{chen2017coherent} proposed an end-to-end network for online video style transfer, which generates temporally coherent stylized video sequences in near real-time. Gao et al.~\cite{gao2020fast} design a multi-instance normalization block to learn different style examples and a ConvLSTM module to encourage temporal consistency. Deng et al.~\cite{deng:2020:arbitrary} proposed a multi-channel correlation network to fuse exemplar style features and input content features for efficient style transfer. Wang et al. ~\cite{wang2020consistent} use relaxation and regularization to consistent video style transfer. However, these methods rely on neural networks for online video processing, which requires more computing resources and processing time, making it difficult to achieve fast video-style transfer.

3D lookup tables (LUTs) are a high-efficiency method typically used for color correction, video enhancement, and retouching~\cite{karaimer2016software,zeng2020learning}. However, creating a LUT is when an experienced expert adjusts various parameters, such as exposure and saturation, with an input image to generate an output image. The input and output images are then mapped to create the LUT. This process requires an expert's expertise and is time-consuming. Once tuned, the LUTs are fixed and not adaptable to different scenes~\cite{zeng2020learning}. Some recent studies have used neural networks to generate LUTs~\cite{zeng2020learning,yang2022adaint,yang2022seplut,liu20224d,zhang2022clut}, and some progress has been made. Zeng et al.~\cite{zeng2020learning} proposed learning image-adaptive 3-dimensional lookup tables. Zhang et al.~\cite{zhang2022clut} proposed the method of adaptively compressed representations for 3D LUT. However, these methods can only enhance the input image and cannot achieve photorealistic style transfer with a pair of content and style images.

In this paper, we propose a method for generating style 3D lookup tables using a neural network, which enables the generation of stylized LUTs for any style and content images. We apply these stylized LUTs to video style transfer, significantly improving the processing speed while maintaining the output quality. Even processing 8K video takes less than 2ms per frame. Our neural LUT network and base LUT are first trained on a large-scale image dataset for photorealistic style transfer. Then, we perform test-time training on specific video content keyframes and style images to obtain the stylized LUT particular to that content video and style image. Finally, we apply the stylized LUT to achieve a photorealistic style transfer for the video. We have implemented an automated method for generating stylized LUTs for arbitrary style, which can theoretically be applied to any video. Leveraging the fast processing capabilities of LUTs, we can achieve super-fast video-style transfer with high-quality output.
In a nutshell, our main contributions are as follows:
\begin{itemize}   
	\item We propose a method to generate stylized LUT based on a neural network. The stylized LUT can achieve ultra-fast video photorealistic style transfer and process 8K resolution video in real time.
	
	\item We introduce test-time training to fine-tune arbitrary style images and content images and realize the LUT generation of photorealistic style transfer in seconds.
	
	\item Experiments demonstrate that our method can not only realize the transfer of photorealistic video style but also maintain high visual quality and consistency.
\end{itemize}
\begin{figure*}[htbp]
	\centering
	\includegraphics[width=0.95\linewidth]{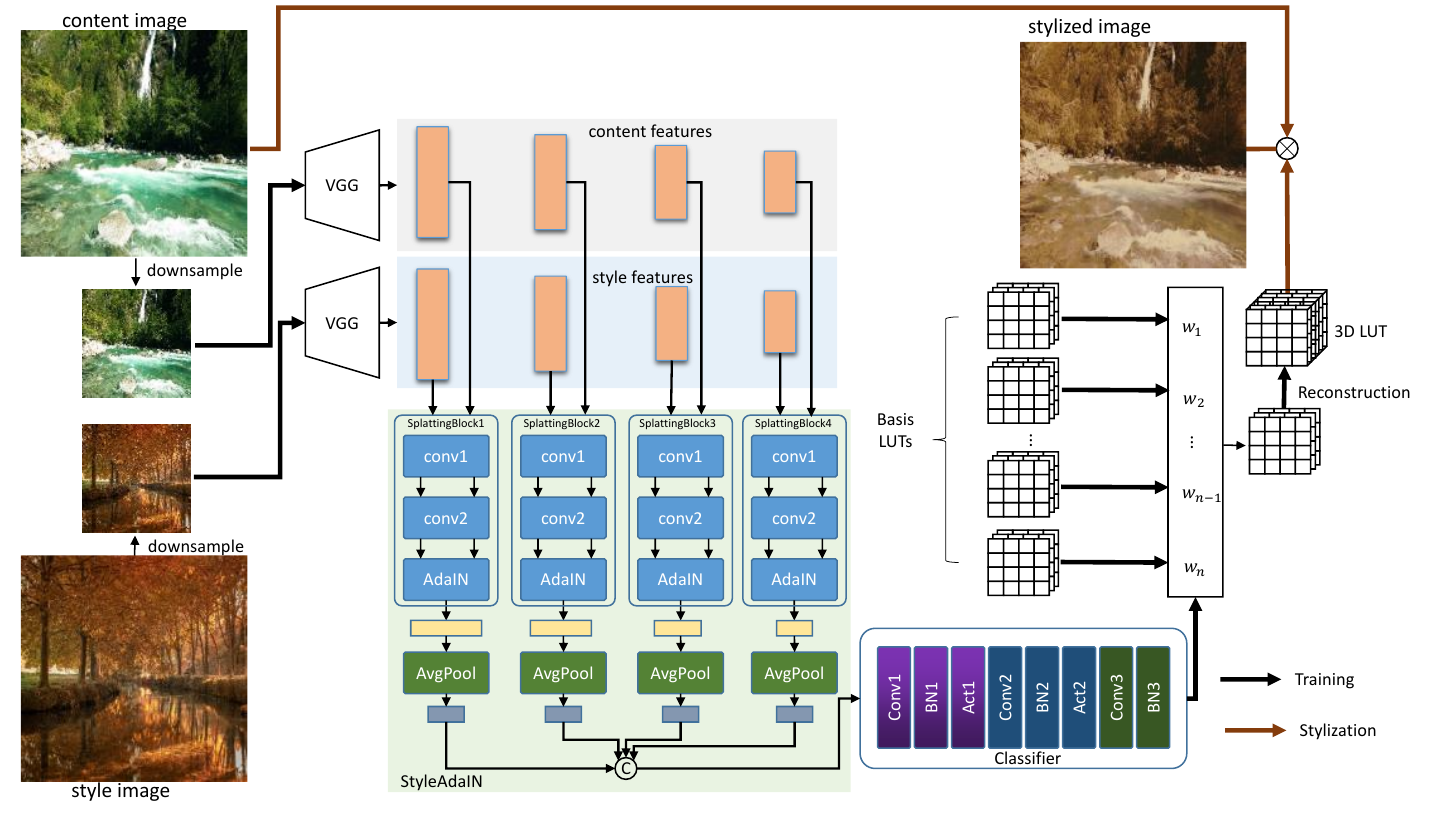}
	\caption{\textbf{The structure of neural LUT network.} In our network structure, we first extract the features of a pair of down-sampled the style image and the content image by the pre-trained VGG model. Then, further feature extraction and fusion of content and style features of different scales are performed, respectively. Finally, a classifier is used to predict the weight of the linear combination of basis LUTs. The parameters of the basis LUTs are learned simultaneously with the network. Only the content image is needed as input during the stylization process, and the reconstructed 3D LUT is used for color query and interpolation to output the stylized image. }
	\label{fig:nlutframework}
	\vspace{-5mm}
\end{figure*}
\section{Related Work}
\label{sec:relate}
\noindent \textbf{Image and video style transfer.}
Artistic style transfer and photorealistic style transfer are two sub-areas of style transfer. They all expect to use the style image as a reference to transform the content image into a stylized image similar to the style image but retain the content in the content image as much as possible~\cite{gatys2015neural,huang2017arbitrary,luan2017deep,li2018closed,yoo2019photorealistic,xia2020joint,qiao2021efficient,kolkin2019style,yoo2021sinir,kim2022deep,chiu2022pca}. Generally, the images created by artistic style transfer are more abstract, so the texture and color of the stylized image are different from the original content image. On the other hand, photorealistic style transfer must retain the texture information of the original content image and only change its color to achieve a photorealistic effect.

Some early methods, such as the histogram of the linear filter response and nonparametric sampling, rely on low-level statistical data and cannot obtain the semantic structure. Gatys et al.~\cite{gatys2016image} introduced neural network deep statistical features for statistics and proposed using a gram matrix for constraints, which can have certain semantic information. To achieve fast stylization, Avatar~\cite{sheng2018avatar} and AdaIN~\cite{huang2017arbitrary} developed feed-forward neural networks. However, these methods cannot realize the transfer of photorealistic style. Although the input image is a photograph, it is still an exhibition display of a painting.

DPST~\cite{luan2017deep} restricts the transformation from input to output to local affine transformation in color space to realize photorealistic style transfer. To improve efficiency, PhotoWCT~\cite{li2018closed} and WCT$^2$\cite{yoo2019photorealistic} were proposed. An end-to-end model for photorealistic style transfer that is both fast and inherently generates photorealistic results has been proposed by  Xia et al.~\cite{xia2020joint}. Qiao et al. ~\cite{qiao2021efficient} proposed Style-Corpus Constrained Learning to relieve the unrealistic artifacts and heavy computational cost issues. However, these methods are all for images. When applied to video style transfer, color flickering usually occurs. The method based on optical flow or temporal constraints is applied to video style transfer to make the stylized video not flicker. The method proposed by MCCNet~\cite{deng:2020:arbitrary}  can be trained to fuse the style features of exemplar and input content features to achieve effective style conversion and obtain consistent results. ReReVST~\cite{wang2020consistent} is used to achieve the consistency of stylized video by jointly considering the intrinsic properties of stylization and temporal consistency. However, these methods need to process video through the neural network during stylization, so they require high computing resources and cannot achieve efficient processing.

\noindent \textbf{LUTs for video processing.} 3D lookup tables (LUTs) are ultrafast operators for color mapping and are widely used to improve the quality of digital images for color correction, video enhancement, and retouching~\cite{karaimer2016software,zeng2020learning}. Recently, many methods based on neural networks to generate LUTs have been proposed. Image-Adaptive-3DLUT ~\cite{zeng2020learning} use a lightweight CNN weight predictor to predict the learned several bases 3D LUTs weights. An adaptive 3D LUT could be generated for each input image by fusing the basis of 3D LUTs according to the image content. AdaINT ~\cite{yang2022adaint} proposed adaptive intervals learning to achieve a more flexible sampling point allocation by adapting the non-uniform sampling intervals in the 3D color space. SepLUT~\cite{yang2022seplut} separates a single color transform into a cascade of component-independent and component-correlated sub-transforms for real-time image enhancement. 4D LUT~\cite{liu20224d} proposed a learnable context-aware 4-dimensional lookup table that achieves content-dependent enhancement of different contents in each image via adaptive learning of photo context. CLUT~\cite{zhang2022clut} analyzed the inherent compressibility of 3D LUT and proposed an effective compressed representation of 3D LUT, which maintains the powerful mapping capability of 3D LUT. These methods explore using a neural network to generate LUT, show the powerful ability of LUT in image and video processing, and to some extent, replace experts to generate LUT manually. However, these methods mainly focus on using the LUT generated by the neural network for image enhancement and cannot achieve the task of photorealistic style transfer of the content image according to the style image. We have thoroughly studied the method of LUT generation by the neural network and realized the photorealistic style transfer of content image for arbitrary style image as a reference. At the same time, the generated LUT is applied to the video to achieve highly efficient video processing.
\begin{figure*}[htbp]
	\centering
	\includegraphics[width=0.95\linewidth]{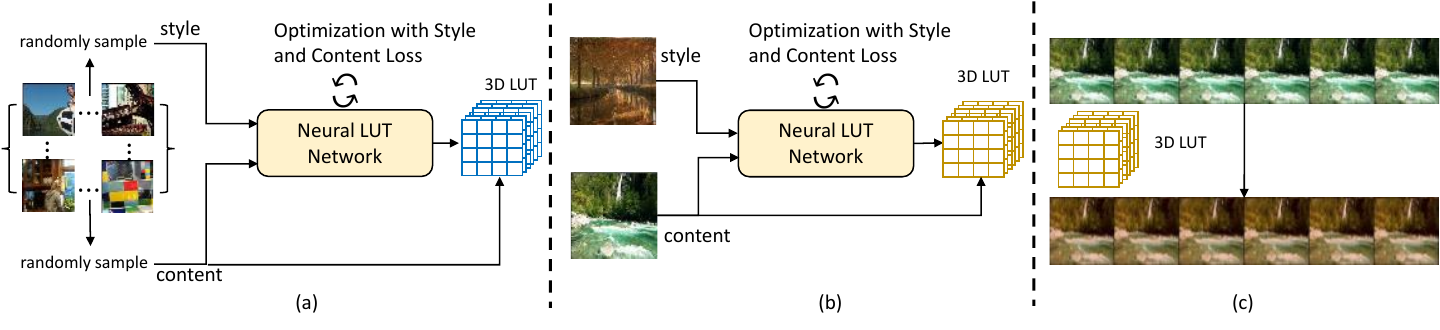}
	\caption{\textbf{Overview of neural-based 3D lookup tables for video photorealistic style transfer.}	(a). First, we randomly sample the style image and content image on a large-scale image dataset to train a neural LUT Network and the initial 3D LUT; (b). Then we use the trained neural LUT network and 3D LUT to conduct tuning training on the content image from the specified video and the specified style image to generate an optimal style 3D LUT; (c). Finally, we use the generated style 3D LUT to interpolate and look up the colors in the content video to get the stylized video.}
	\label{fig:OverviewNLUT}
	\vspace{-4mm}
\end{figure*}

\section{Our approach}
The overview of neural-based 3D lookup tables for video photorealistic style transfer has been shown in Fig.~\ref{fig:OverviewNLUT}. We first train our neural LUT network on large datasets to generate stylized 3D LUTs. Then, we perform test-time training on specific content and style images to create specific stylized 3D LUT. Finally, the generated stylized 3D LUT is used for video photorealistic style transfer. The structure of our neural LUT network is shown in Fig.~\ref{fig:nlutframework}. Our method can generate stylized 3D LUT of arbitrary style images and content images and be more efficient and photorealistic when applied to video processing.

\subsection{LUT for image processing} \label{sec:SceneGeometricReconstruction}
The 3D LUT is typically represented by a three-dimensional array denoted as  $\operatorname{\bm{\phi}} \in \mathbb{R}^{3\times{D_{x}}\times{D_{y}}\times{D_{z}}}$. Under normal conditions $D_{x}$, $D_{y}$, $D_{z}$ are equal to $D$. This array represents complex color mapping relationships and is widely used in image retouching applications. $\operatorname{\bm{\phi}}$ can be split into three sub-tables corresponding to different channels by $\operatorname{[\bm{\phi}}^{c}]_{c \in {r,g,b}}$ where $\operatorname{\bm{\phi}^{c}} \in \mathbb{R}^{{D_{x}}\times{D_{y}}\times{D_{z}}}$. 3D LUT discretizes each dimension of the RGB color space into ${D}$ bins, resulting in $\operatorname{D^{3}}$ discrete colors denoted by $\{(i,j,k)\}_{i,j,k=1,...,D}$, and stores the corresponding mapped color of each of them as $\operatorname{(\bm{\phi}^{r}_{[i,j,k]},\bm{\phi}^{g}_{[i,j,k]},\bm{\phi}^{b}_{[i,j,k]})}$, where  $\operatorname{\bm{\phi}^{r}_{[i,j,k]}}$ represents the element with $\operatorname{[i,j,k]}$ coordinate in  $\operatorname{\bm{\phi}^{c}}$. Take the input image as $I$ and the output image as $O$, the process of generating $O$ is complex and usually involves sampling the color space, calculating the output values at each sampling point, and using a 3D interpolation algorithm to interpolate the unsampled points in the color space, thus obtaining a continuous and smooth mapping function. The color transformation process of 3D LUT can be denoted as follows:
\begin{equation} \label{eq:LUTImage}
	{O} =\bm{\phi}(I) ~ ,
\end{equation}

CLUT~\cite{zhang2022clut} employs learning adaptively compressed representations of 3D LUTs for image enhancement. It compressed $\operatorname{\bm{\phi}^{c}}$ to $\operatorname{\bm{\psi}^{c}}$ by reducing ${{D}_{x}}$ to $S$ and ${{D}_{y}}\times{D_{z}}$ to $W$ with linear spatial transformation, in which two utilized transformation matrices denoted by $\operatorname{\bm{M}=\{\bm{M}_{s} \in\mathbb{R}^{{D}\times{S}},\bm{M}_{w} \in\mathbb{R}^{{W}\times{D^{2}}}\}}$ server as the suitable compression schemes for different dimensions. In our work, the default value of $S$ is $32$, and the default value of $W$ also is $32$. The 3D LUT reconstruction process from CLUT can be conducted as follows:
\begin{equation} \label{eq:photo_loss}
	\bm{\phi}=f(\bm{\psi},\bm{M})=h([\bm{M}_{s}\bm{\psi}^{r}\bm{M}_{w}, \bm{M}_{s}\bm{\psi}^{g}\bm{M}_{w},\bm{M}_{s}\bm{\psi}^{b}\bm{M}_{w} ]) ~ ,
\end{equation}
where $\operatorname{h}$ denotes a simple reshape procedure from $\operatorname{\mathbb{R}^{{3}\times D \times{D^{2}}}}$ to $\operatorname{\mathbb{R}^{{3}\times D\times D\times D}}$. Thus Eq. ~\ref{eq:LUTImage} can be further expressed as:
\begin{equation} \label{eq:LUTImage}
	{O} =\bm{\phi}(I)= f(\bm{\psi},\bm{M})(I)~ ,
\end{equation}


\subsection{Neural LUT network for 3D LUT generation} \label{sec:2d_style}
We have designed a neural network to generate stylized 3D LUTs, called neural LUT network, as shown in Fig.~\ref{fig:nlutframework}. The input of the neural LUT network is a style image and a content image, and the output is a corresponding 3D LUT, which can achieve photorealistic style transfer accordingly.

We downsample the full-size style image  ${I_f^s}$ and content image ${I_f^c}$ to obtain the corresponding low-resolution style image ${I_f^c}$ and content image ${I_l^c}$ respectively, thus reducing the GPU memory requirements and computation time consumption. Then we get the multi-scale style features  $\{F_j^s\}_{1 \leq j \leq  4}$ and content features $\{F_j^c\}_{1 \leq j \leq  4}$ by pre-trained VGG network~\cite{simonyan2014very} for feature extraction of style image and content image respectively.

Furthermore, we perform convolutional feature extraction on style and content features at different feature scales, respectively, and fuse them by adaptive instance normalization (AdaIN)~\cite{huang2017arbitrary}. Finally, the features after each AdaIN operation are adaptively pooled (AvgPool)~\cite{lin2013network} to obtain features of the same dimension. These same dimensional features are then concatenated as input to the $classifier$ for the N-base linear weight prediction of CLUTs.

In our work, we use a linear combination of the $\operatorname{N}$ base of CLUTs, which cover different color effects that are required in different scenes, to denote the final CLUT:
\begin{equation} \label{eq:photo_loss}
	\bm{\psi} = \sum_{n=1}^{N}\bm{w}_{i}\bm{\psi}_{i}  ~ ,
\end{equation}
where $\bm{\psi}_{i}$ is one of the $N$ basis CLUTs and $\bm{w}_{i}$ is its corresponding linear combination weight parameter. It is important to note that the parameters of these N-base CLUTs are learned during network training.

Thus, for the neural LUT network, it can be represented as:
\begin{equation} \label{eq:photo_loss}
	\bm{W} =\{\bm{w}_{i}\}_{1 \leq i \leq  N}=f_{NLUT}({I_f^s},{I_f^c}) ~ ,
\end{equation}

Specifically, during the inference phase, a stylized CLUT $\bm{\psi}_{sed}$ can be obtained based on the input style image  ${I_f^s}$ and content image ${I_f^c}$. Thus, Eq. \ref{eq:LUTImage} can be rewritten as:
\begin{equation} \label{eq:LUTImage}
	I_{sed} =\bm{\phi}_{sed}({I_f^c})= f(\bm{\psi}_{sed},\bm{M})({I_f^c})~ ,
\end{equation}

\subsection{Loss function} \label{sec:SceneGeometricReconstruction}
Similar to AdaIN~\cite{huang2017arbitrary}, we define our content loss ${L_c}$ and style loss ${L_s}$.  We use the pre-trained VGG model to extract the feature maps of the content, style, and stylized images, respectively. For the style loss, defined as:
\begin{equation} \label{eq:style_loss}
	{L_{s}} = \sum_{i=1}^{4}||\mu{(F^{i}_{s})}-\mu{(F^{i}_{sed})}||_{2}+\\
	||\sigma {(F^{i}_{s})}-\sigma {(F^{i}_{sed})}||_{2}  ~ ,
\end{equation}
where $F^{i}$ denotes the $i$-th layer of the feature map extracted by the pre-trained VGG model, $\mu()$ and $\sigma()$ denote a layer in VGG used to compute the means and variances, respectively.
For content loss, we only use the last layer of the feature maps extracted by the pre-trained VGG model :
\begin{equation} \label{eq:style_loss}
	{L_{c}} = ||{(F^{4}_{s})}-{(F^{4}_{sed})}||_{2} ~ ,
\end{equation}

To ensure the generated 3D LUT has the smooth color mapping and surfaces, we follow ~\cite{zeng2020learning} smooth and monotonicity regularization constraints to define regularization losses $R_{s}$, $R_{m}$. Our overall loss is the weighted summation of style loss ${L_{s}}$, content loss ${L_{c}}$ and regularization loss:
\begin{equation} \label{eq:style_loss}
	{L} = \lambda_{s}{L_{s}}+ \lambda_{c}{L_{c}}+\lambda^{r}_{s}{R_{s}}+\lambda^{r}_{m}{R_{m}}~ ,
\end{equation}
where $\lambda_{s}$, $\lambda_{c}$, $\lambda^{r}_{s}$ and $\lambda^{r}_{m}$ are hyper-parameters controlling weights of their corresponding loss terms.

\subsection{Training strategies} \label{sec:2d_style}

As shown in Fig.~\ref{fig:OverviewNLUT}, we first train the neural LUT network and the base CLUT on a large-scale image dataset MS-COCO~\cite{2014Microsoft}, where both style images and content images are generated by random sampling, and then trained by constraining the style loss and content loss. Then we extract the keyframe from the specific video as the content image and use the specified style image for model fine-tuning. Since the specific style image and the content video need to be re-trained in order to get the corresponding style LUT during the test, we refer to this stage as the ``test-time training''. We use the fine-tuning network to generate the style 3D LUT. Finally, we use the generated style 3D LUT to interpolate and lookup the colors in the content video to get the stylized video. 
\begin{figure*}[htbp]
	\centering
	\includegraphics[width=0.85\linewidth]{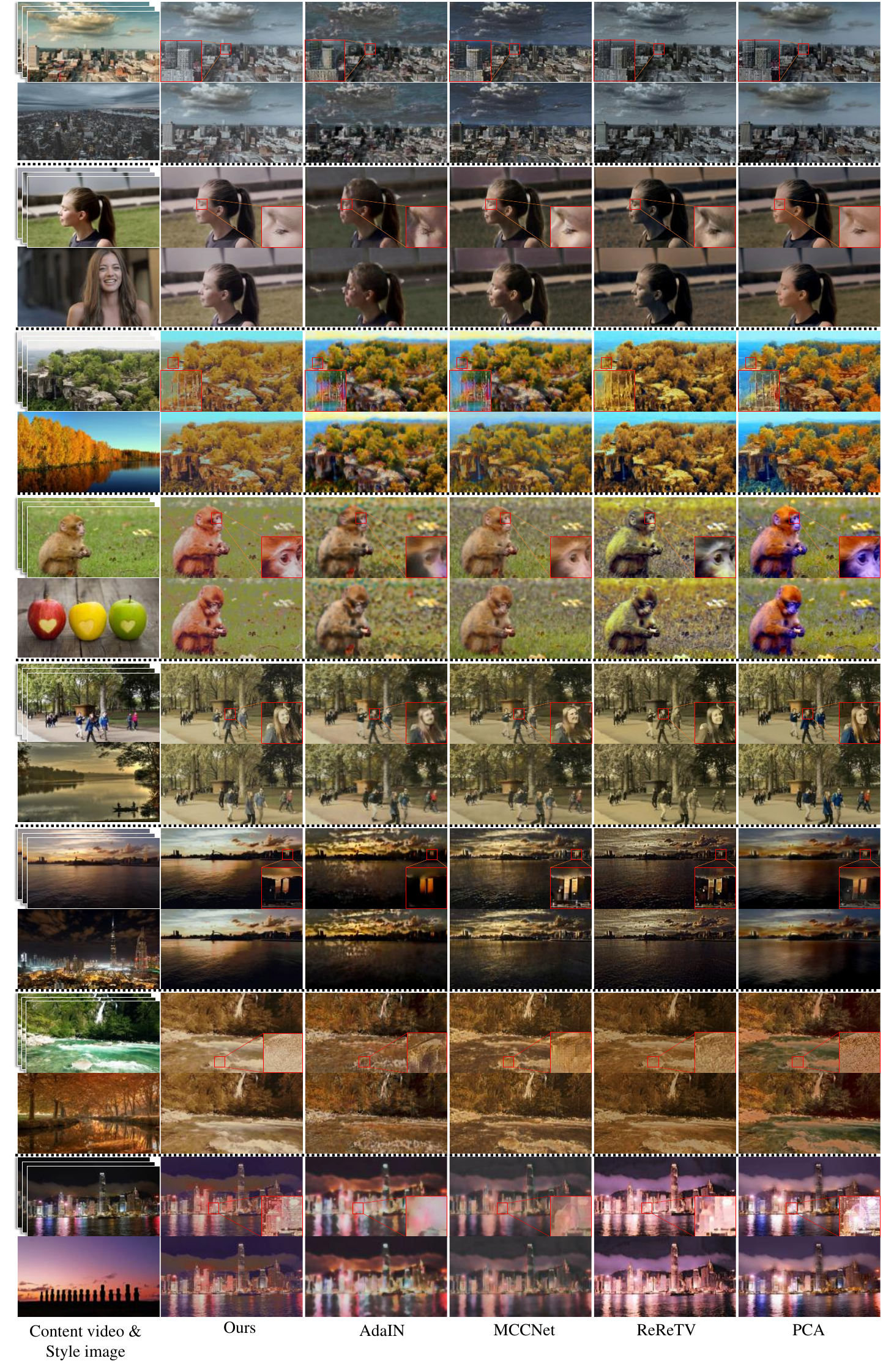}
	\caption{\textbf{Qualitative comparisons with photorealistic style videos.} We compare the stylized results of 8 videos collected from  $pixabay$ and $pexels$. Our method stylizes scenes with more precise details and competitive stylization quality.}
	\label{fig:photorealistivideosc}
\end{figure*}
\begin{figure*}[htbp]
	\centering
	\includegraphics[width=0.95\linewidth]{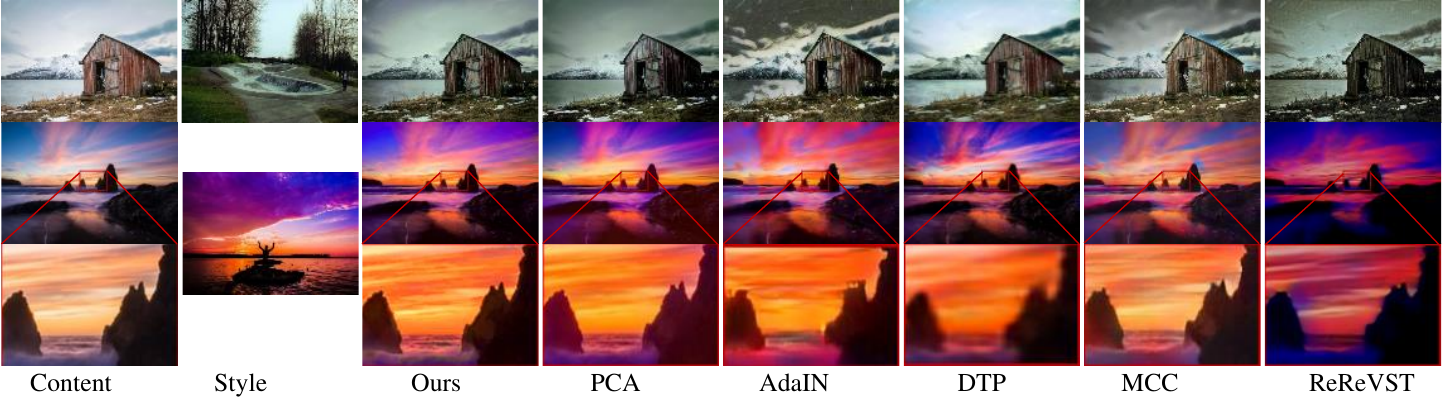}
	\caption{\textbf{Qualitative comparisons with images}. Compared to others, our model generates realistic results while successfully transferring both global and local style
		information and preserving structure.}
	\label{fig:img_comp}
\end{figure*}

\section{Experiments}
\label{sec:experiment}

We have done the qualitative and quantitative evaluation tests for our method and also comparisons with the state-of-the-art stylization methods for image and video, respectively. In our training stage, we use the Adam optimizer with a batch size of 6 to optimize the neural LUT network and basis CLUTs for 350k iterations; in our fine-tuning training stage, we use the Adam~\cite{kingma2014adam} optimizer with a batch size of 8 to optimize the neural LUT network for about 20 iterations (less in 30 seconds). The learning rate is $10^{-4}$. We use images in the MS-COCO~\cite{2014Microsoft} as the reference style images in the training stage. The pair of content and style images are bilinearly resized to the size of $256 \times 256$ in our training stage. All experiments are performed on a single NVIDIA TITAN RTX GPU. The videos for testing are collected from $pixabay$\footnote{https://pixabay.com/videos} and $pexels$\footnote{https://www.pexels.com/}, and the length of each video has 6 seconds with 25 fps.

\subsection{Qualitative results}

\noindent\textbf{Photorealistic style transfers with videos.} We compare our method
with image-based approaches (AdaIN~\cite{huang2017arbitrary} and PCA~\cite{chiu2022pca}), as well as video-based approaches (MCCNet~\cite{deng:2020:arbitrary}, ReReVST~\cite{wang2020consistent}), and report in Fig.~\ref{fig:photorealistivideosc}. Obviously, our approach not only achieves a photorealistic style transfer of the video in terms of visual effects but also preserves the details of the video content better than other methods. For example, in the video of ``city", our method can see the texture of distant buildings more clearly than other methods. This is due to the fact that 3D LUT inherently has migration to color, and we have implemented 3D LUT generation using neural networks.

\noindent\textbf{Photorealistic style transfers with images.} In Fig.~\ref{fig:img_comp}, we qualitatively compare the photorealistic style transfer results generated by
PCA~\cite{chiu2022pca}, AdaIN~\cite{huang2017arbitrary}, DTP~\cite{kim2022deep}, MCCNet~\cite{deng:2020:arbitrary} and ReReVST~\cite{wang2020consistent}. The resolution of the content image we tested is $1000\times667$. Since DTP~\cite{kim2022deep} needs to resize the image to $256\times256$ when inputting to the network, the results we obtained during the test were upsampled back to the original resolution. Our results not only realize the transfer of photorealistic style but also hardly lose the quality of the content image. AdaIN, DTP, MCC, and ReReVST all lose quality. For example, the snow mountain behind the house in AdaIN is almost more artistic, while the mountain in the distance in ReReVST is almost invisible. Compared to PCA, we don't seem to have much difference in vision, but we have speed and memory consumption advantages in video stylization.

\subsection{Quantitative results}
\noindent\textbf{Consistency measurement.} We measure the short and long-term consistency using the warped Learned Perceptual Image Patch Similarity (LPIPS) metric~\cite{zhang2018unreasonable}. We use the measurement implemented from ~\cite{lai2018learning}. The consistency score formulates as follows: 
\begin{equation}
	E(O_i, O_j) = LPIPS(O_i, M_{i,j}, W_{i,j}(O_j))
\end{equation}
where $W$ is the warping function and $M$ is the warping mask. When calculating the average distance across spatial dimensions in ~\cite{zhang2018unreasonable}, only pixels within the mask $M_{i,j}$ are taken. We compute the evaluation values on 8 videos from $pixabay$ and $pexels$. We use interval of 5 frames ($O_{i},O_{i+5}$) and 35 ($O_{i},O_{i+35}$) for short and long-range consistency calculation. The short- and long-range consistency comparisons are shown in Tab.~\ref{tab:short} and Tab.~\ref{tab:long}, respectively. Our method outperforms other methods by a significant margin.

\begin{table}
	\centering
	\caption{\textbf{Short-range consistency.} We compare the short-range consistency using warping error($\downarrow$). {\textbf{Best}} {results are highlighted.}}\label{tab:short}
	\resizebox{\columnwidth}{!}{
		\begin{tabular}{l|cccccc|c}
			\hline
			Method	&				city		&				girl		&				kelly		&				monkey		&				pedestrian	&				sunset		&				Average		\\
			\hline																													
			AdaIN	&				0.0019 		&				0.0004 		&				0.0044 		&				0.0012 		&				0.0023 		&				0.0020 		&				0.0025 		\\
			MCCNet	&				0.0033 		&				0.0003 		&				0.0024 		&				0.0007 		&				0.0019 		&				0.0015 		&				0.0016 		\\
			ReReVST	&				0.0010 		&				0.0002 		&				0.0015 		&				0.0014 		&	\textbf{	0.0013 	}	&	\textbf{	0.0009 	}	&				0.0013 		\\
			PCA		&				0.0011 		&				0.0002 		&				0.0010 		&				0.0009 		&				0.0021 		&				0.0022 		&				0.0015 		\\
			\hline																													
			Ours	&	\textbf{	0.0008 	}	&	\textbf{	0.0001 	}	&	\textbf{	0.0008 	}	&	\textbf{	0.0006 	}	&				0.0015 		&				0.0015 		&	\textbf{	0.0011 	}	\\
			\hline
		\end{tabular}
	}
	\vspace{-2mm}
\end{table}

\begin{table}
	\centering
	\caption{\textbf{Long-range consistency.} We compare the long-range consistency using warping error($\downarrow$).  {\textbf{Best}} {results are highlighted.}}\label{tab:long}
	\resizebox{\columnwidth}{!}{
		\begin{tabular}{l|cccccc|c}
			\hline
			Method	&				city		&		girl		&		kelly		&		monkey		&		pedestrian		&		sunset		&		Average		\\
			\hline																													
			AdaIN	&				0.0047 		&		0.0029 		&		0.0096 		&		0.0028 		&		0.0057 		&		0.0025 		&		0.0052 		\\
			MCCNet	&				0.0075 		&		0.0046 		&		0.0055 		&		0.0020 		&		0.0045 		&		0.0015 		&		0.0034 		\\
			ReReVST	&				0.0028 		&		0.0032 		&		0.0078 		&		0.0032 		&	\textbf{	0.0032 	}	&		0.0027 		&		0.0042 		\\
			PCA		&				0.0062 		&		0.0039 		&		0.0077 		&		0.0102 		&		0.0118 		&		0.0017 		&		0.0078 		\\
			\hline																													
			Ours	&	\textbf{	0.0030 	}	&	\textbf{	0.0018 	}	&	\textbf{	0.0044 	}	&	\textbf{	0.0018 	}	&		0.0035 		&		\textbf{0.0005} 		&	\textbf{	0.0026 	}	\\
			\hline

	\end{tabular}}
	
\end{table}
\begin{figure}[htbp]
	\vspace{1mm}
	\centering
	\includegraphics[width=0.85\linewidth]{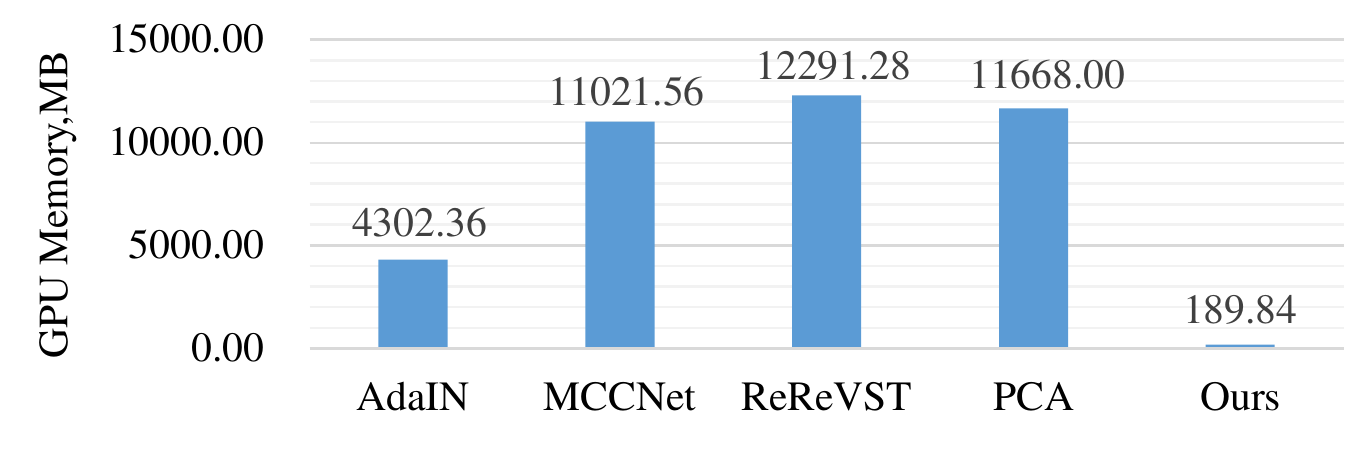}
	\caption{\textbf{GPU memory usage comparison for 4K video.} }
	\label{fig:GPUm}
\end{figure}
\begin{table}
	\centering
	\caption{\textbf{Efficiency comparison.} We compare the efficiency at various resolutions using ``millisecond per image"($\downarrow$).  {\textbf{Best}} {results are highlighted.}}\label{tab:inferencetime}
	\resizebox{\columnwidth}{!}{
		\begin{tabular}{l|cccccccc}
			\hline
			Method	&		512 		&		HD		&		FHD		&		QHD		&		2000		&		4K		&		5K		&		8K		\\
			\hline																													AdaIN	&		36.28 		&		118.32 		&		255.19 		&		412.26 		&		444.68 		&		1025.05 		&		1832.61 		&		OOM		\\				
			MCCNet	&		63.83 		&		209.81 		&		465.52 		&		807.26 		&		913.65 		&		2045.06 		&		OOM		&		OOM		\\
			
			ReReVST	&		29.58 		&		101.25 		&		221.86 		&		409.54 		&		471.02 		&		980.15 		&		OOM		&		OOM		\\
			PCA	&		65.76 		&		74.50 		&		101.99 		&		186.81 		&		198.48 		&		381.22 		&		669.99 		&		OOM		\\
			\hline																																	
			Ours	&	\textbf{	0.05 	}	&	\textbf{	0.05 	}	&	\textbf{	0.10 	}	&	\textbf{	0.19 	}	&	\textbf{	0.27 	}	&	\textbf{	0.43 	}	&	\textbf{	0.76 	}	&	\textbf{	1.72 	}	\\
			\hline

	\end{tabular}}
	\vspace{-5mm}
\end{table}

\begin{figure}[htbp]
	\centering
	\includegraphics[width=0.95\linewidth]{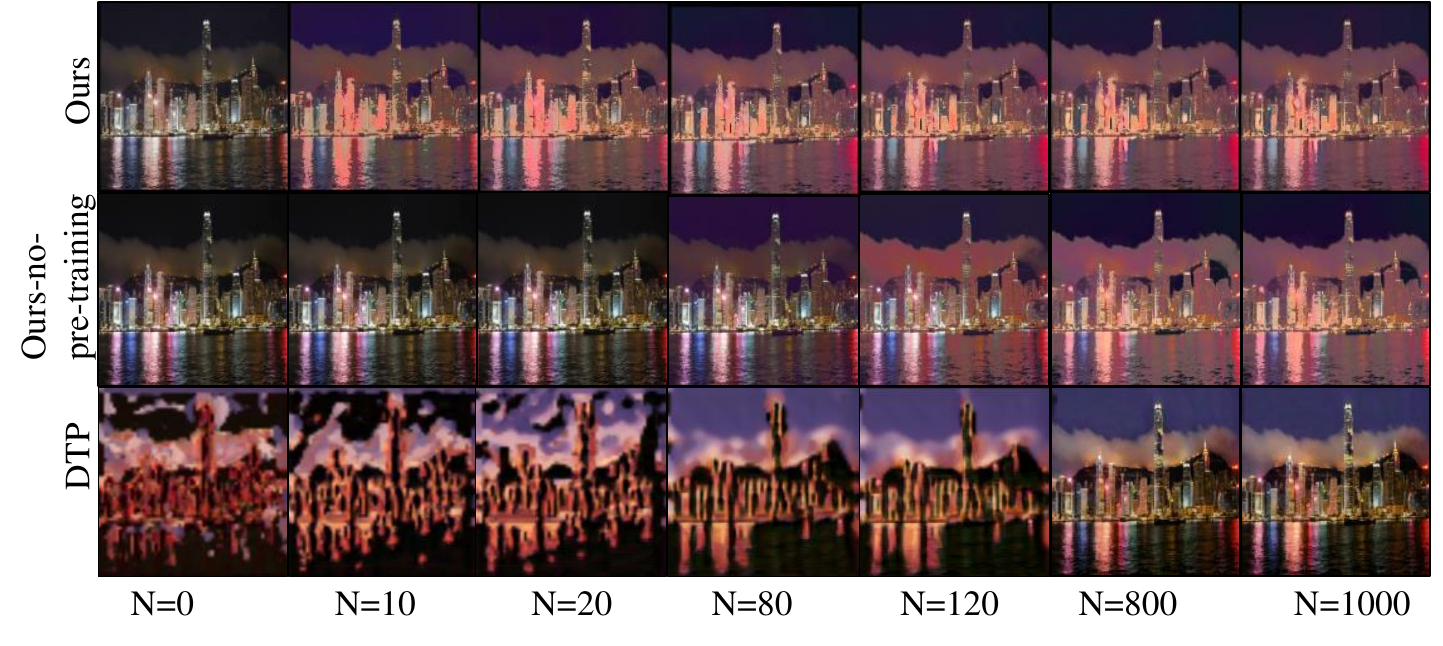}
	\caption{\textbf{Test-time training comparisons with DTP~\cite{kim2022deep}.} The top is the process of our test-time process; the middle is the process of our method without pre-training on a large dataset for tuning; the bottom is the process of DTP~\cite{kim2022deep} test-time training. }
	\label{fig:Testtimetraining}
	\vspace{-4mm}
\end{figure}

\begin{figure}[htbp]
	\vspace{1mm}
	\centering
	\includegraphics[width=0.65\linewidth]{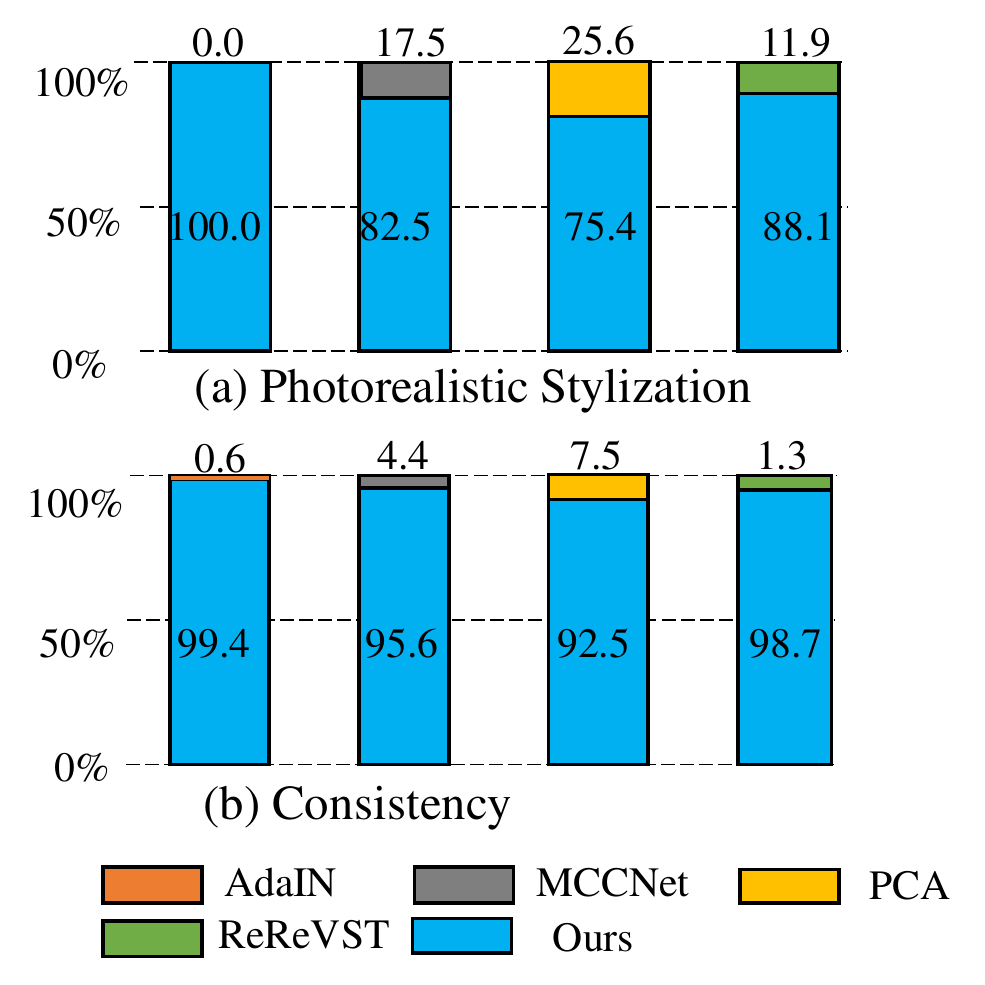}
	\caption{\textbf{User study.}  Our results win more preferences both in the photorealistic stylization and consistency quality.}
	\label{fig:user_study}
	\vspace{-4mm}
\end{figure}
\noindent\textbf{Efficiency comparison.} We compare the efficiency at various resolutions using ``millisecond per image". We have tested on various resolutions: $512\times 512$ ($512$), $1280\times 720$ (HD), $1920\times 1080$ (FHD), $2560\times 1440$ (QHD), $2000\times 2000$ (2000), $3840\times 2160$ (4K), $5120\times 2880$ (5K), and $7680\times 4320$ (8K). We used the official code provided by the comparison method and tested it on the same computer using a single NVIDIA TIATAN RTX GPU with 24GB RAM. As shown in Tab. ~\ref{tab:inferencetime}, our method is extremely fast and is much faster than other methods. Even at $8K$ resolution, it only takes less than $2$ milliseconds. This is because other methods require the forward operation of the neural network when performing the video style transfer. We only need to query the color of the input image and perform trilinear interpolation by the generated stylized 3D LUT, without any convolution operation, when we perform video style transfer.

\noindent\textbf{GPU memory usage comparison.} In Fig.~\ref{fig:GPUm}, we compare the GPU memory occupation of various methods when processing 4K video. Since our method only needs stylized 3DLUT for video processing, the occupation of video memory in the stylized video is also very small compared with other methods.

\noindent\textbf{User study.} A user study is conducted to compare our method's stylization and consistent quality with other state-of-the-art methods. We stylize 8 videos with different methods (AdaIN~\cite{huang2017arbitrary}, MCCNet~\cite{deng:2020:arbitrary}, PCA~\cite{chiu2022pca}, and ReReVST~\cite{wang2020consistent}) and invite 20 participants. First, we showed the participants a style image, the content video, and two stylized videos generated by our method and a compared method. Then we recorded the number of participants' evaluation votes on the photorealistic stylization and consistency of the stylized videos. We collected 1280 (for each method being compared, 20 participants asked questions about 2 aspects of 8 videos) votes for each evaluating indicator and presented the result in Fig.~\ref{fig:user_study} in the form of the boxplot. Our method is outstanding in terms of photorealistic stylization and consistency.

\subsection{Ablation study}

\noindent\textbf{The impact of pre-training for test-time training stage.} The network trained on large datasets has better generalization performance when generating LUTs. The basis LUTs can also better represent different colors. We have verified this idea and compared it to the training process used by the DTP~\cite{kim2022deep}. For a fair comparison, we set the batch size to 2 when training our network and DTP. Fig. \ref{fig:Testtimetraining} compares the training process. The result of our photorealistic style transfer is perfect when the number of iterations reaches $10$. This process takes less than $15$ seconds. Even when we reach the default number of iterations ($20$), it takes no more than $30$ seconds. However, to achieve a comparable style transfer result, DTP needs nearly $1000$ iterations, which takes more than $200$ seconds. Suppose we do not use the pre-training model to initialize the test-time training. In that case, this process may require nearly 80 iterations (as shown in the middle of Fig. \ref{fig:Testtimetraining}), but this is still less than the training iterations of DTP. Therefore, it is effective to use the pre-training of the neural LUT network on a large dataset.

\begin{figure}[htbp]
	\vspace{1mm}
	\centering
	\includegraphics[width=0.95\linewidth]{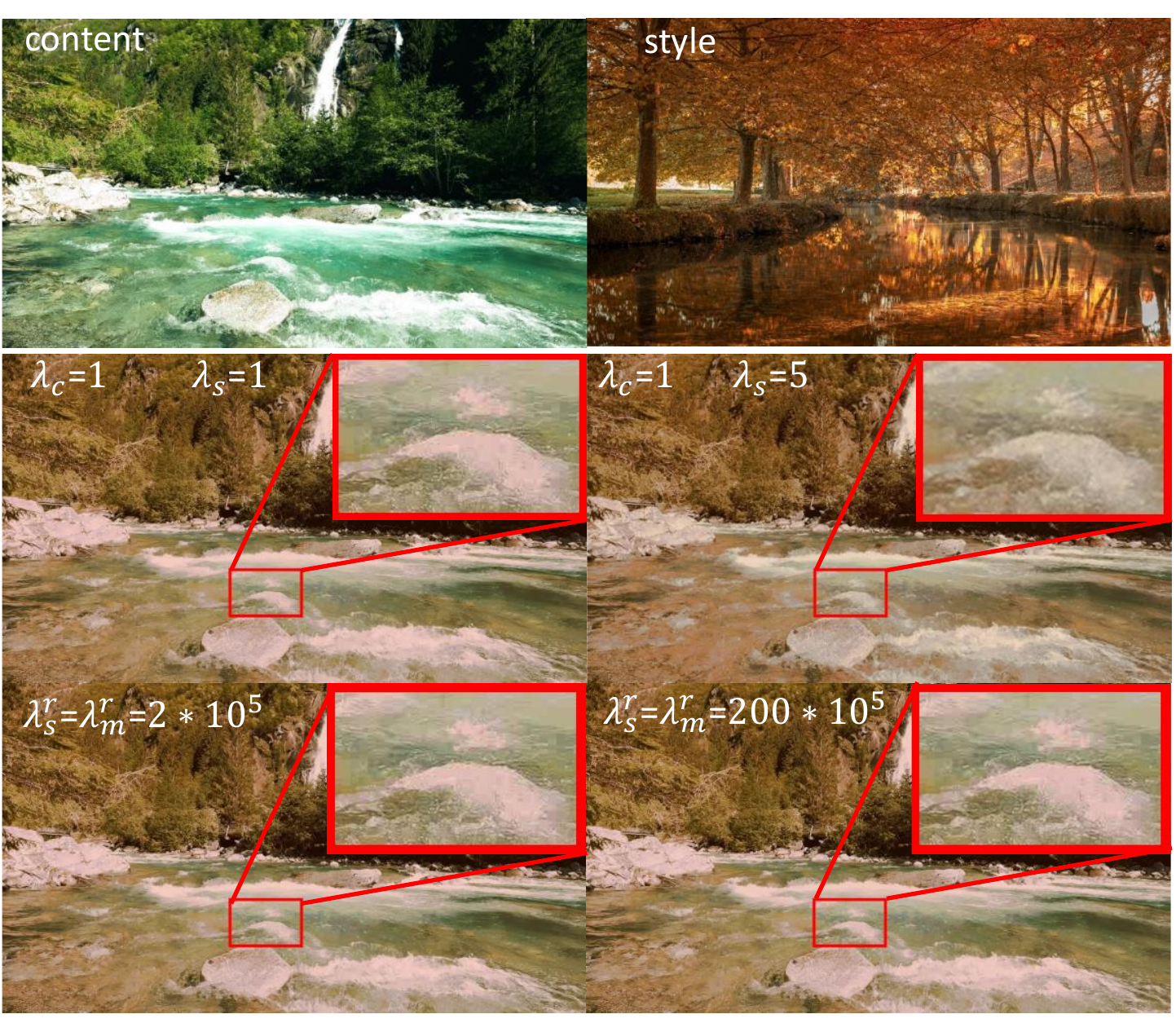}
	\caption{\textbf{The impact of hyper-parameters controlling weights of the loss.} The upper shows the impact of style and content loss controlling weights, and the lower shows the impact of regularization loss controlling weight.}
	\label{fig:ab_loss}
	\vspace{-4mm}
\end{figure}
\noindent\textbf{The impact of hyper-parameters controlling weights of the loss.} The weight of style loss, content loss, and regularization loss affects the result of video stylization. When we fixed the iteration number of test-time training to $20$, we changed different loss weight parameters to experiment. The experimental results are shown in Fig.~\ref{fig:ab_loss}. It can be seen from the figure that the relative weight of style loss and content loss has a great impact on the result: If we fix the weight of regularization loss, the bigger the style loss, the more the stylized video color is biased toward the style image; on the contrary, it prefers the content image. When the weight of style loss and content loss is fixed, the change of regularization loss has little effect on the result. The reason may be that the 3D LUT has converged well due to the regulation loss during the pre-training of the large dataset, and the shorter iteration number has little effect on the results during the test-time training.
\begin{figure}[htbp]
	\vspace{1mm}
	\centering
	\includegraphics[width=0.95\linewidth]{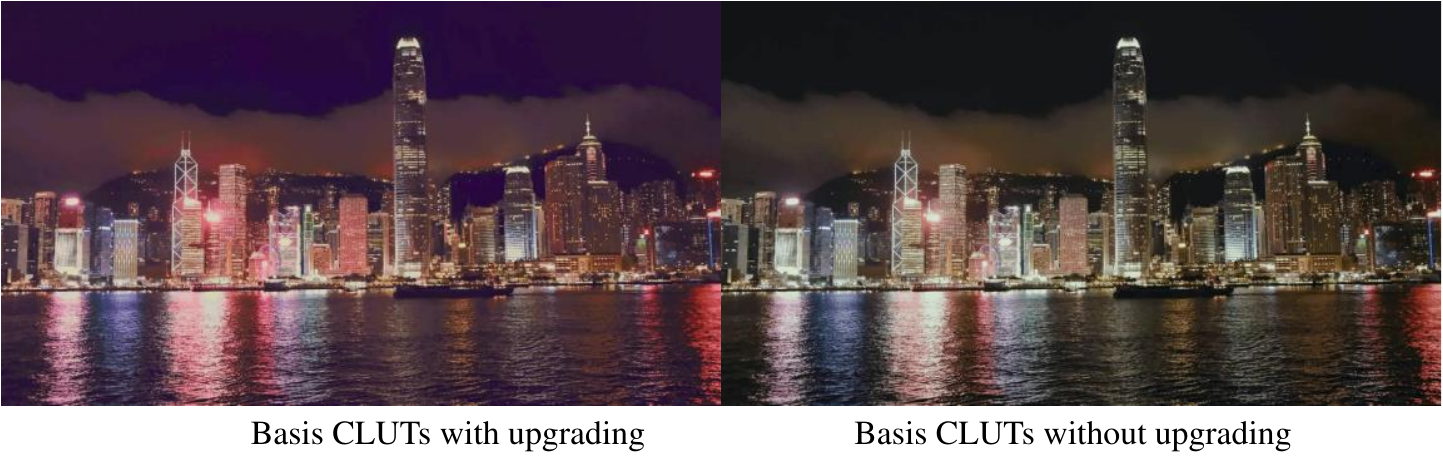}
	\caption{\textbf{The impact of the basis CLUTs upgrading in test-time training stage.} }
	\label{fig:ab_CLUTs}
	\vspace{-4mm}
\end{figure}

\noindent\textbf{The impact of the basis CLUTs upgrading in test-time training stage.} We have a wide range of color representation capabilities because our initial CLUTs parameters are trained from large datasets. Nevertheless, we found in the experiment that in the test-time training stage, if we continue to update the parameters of basis CLUTs, the effect of photorealistic style transfer is better than that of optimizing without updating the parameters of CLUTs. Fig. \ref{fig:ab_CLUTs} compares the two situations with the stylized images generated under the same iterations ($20$) and the same batch size ($2$). The left figure updates the basis CLUTs in test-time training, while the right figure does not. The reason for this may be that the parameters of the basis CLUTs continue to be optimized in the test-time training stage. On the other hand, the reason may be that it is more targeted to continue to optimize the parameters of the basis CLUTs in the test time training stage.

\begin{figure}[htbp]
	\vspace{1mm}
	\centering
	\includegraphics[width=0.95\linewidth]{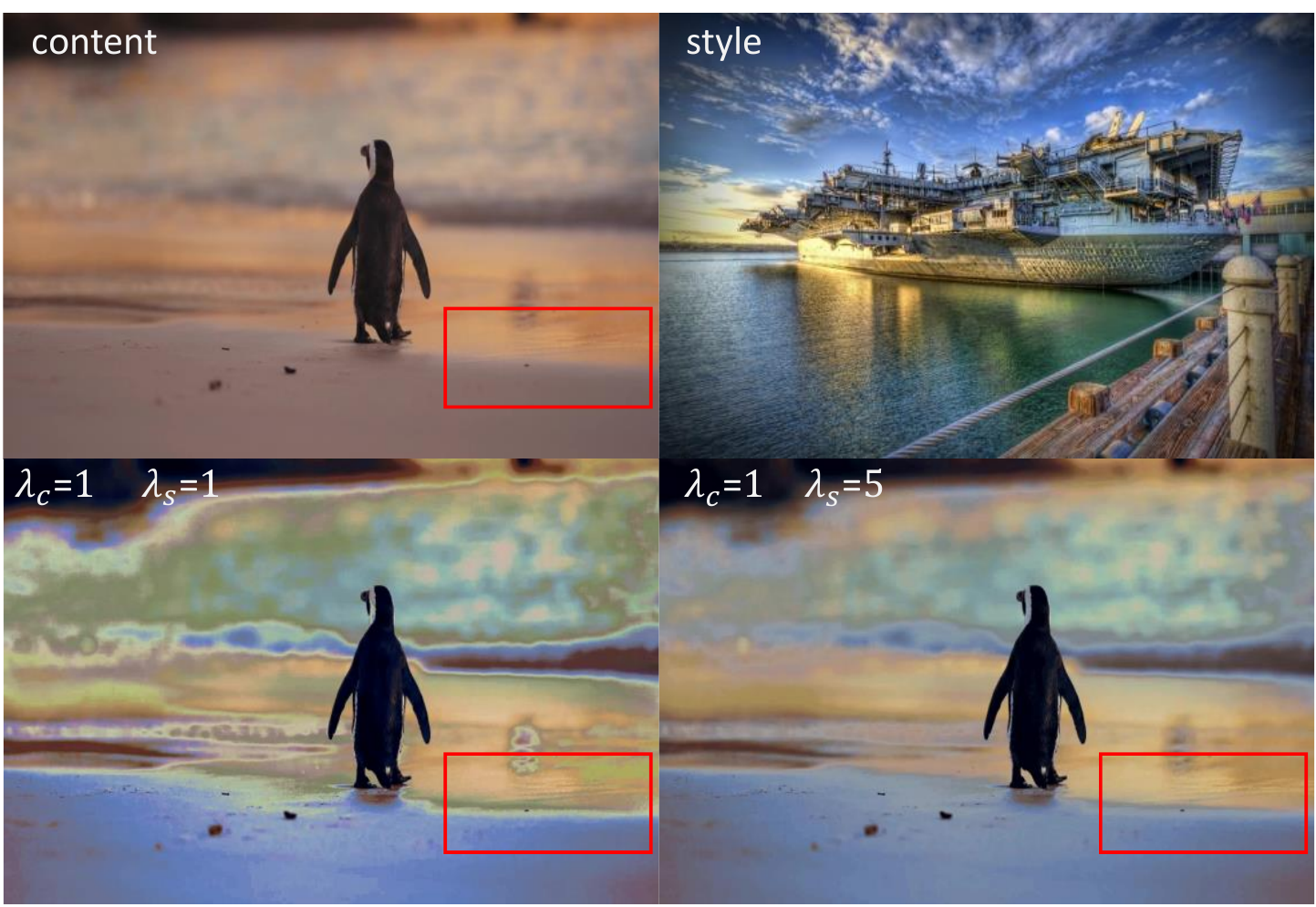}
	\caption{\textbf{Color spots.} Increasing the constraint of content loss during test time training will avoid this defect.}
	\label{fig:limitation}
	\vspace{-4mm}
\end{figure}

\noindent\textbf{Limitations.} LUT will cause color spots to appear in the processing of transition pixels. This defect can occur especially when processing blurred, cloudy, and other scenes. Therefore, our method also has such defects. One way to avoid it is to increase the constraint of content loss during test-time training, which will make the learned style LUT more inclined to content image, thus reducing the possibility of color spots. Fig.\ref{fig:limitation} shows the defects and the effect of the workaround.
\section{Conclusion}
We present a super fast photorealistic style transfer method for video. We build a neural network to generate a stylized 3D LUT. The goal is to realize fast photorealistic style transfer for video. Specifically, we train the neural network that produces 3D LUT on a large dataset and then fine-tune it in test-time training to generate a stylized 3D LUT of a specific style image and video content. Although our method needs fine-tuning when used, it is more effective than other methods and is super fast in video processing. For example, it can process 8K video in less than 2 milliseconds. In the future, we will explore ways to generate 3D LUTs in arbitrary styles even more quickly.

\section*{Acknowledgements}
We sincerely appreciate all participants in the user study.

\clearpage
\balance
\bibliographystyle{ACM-Reference-Format}
\bibliography{sample-sigconf}

\clearpage
\appendix

\section{Detailed configuration of neural LUT network }

Tab.~\ref{tab:StyleAdaIN}, and~\ref{tab:Clasifier} are the detailed configurations of the neural networks used in our framework of neural LUT network which shown in Fig.\ref{fig:nlutframework}. Relevant abbreviations are defined as follows: $\operatorname{OP}$ refers to  $\operatorname{Operation}$, $\operatorname{IN}$ refers to  number of the $\operatorname{Input}$ channels of the features,  $\operatorname{OUT}$ refers to  number of the $\operatorname{Output}$ channels of the features and $\operatorname{ACT}$ refers to the $\operatorname{Activation}$ function. In Tab.~\ref{tab:StyleAdaIN}, $\operatorname{AdaIN}$ is the Adaptive Instance Normalization defined in ~\cite{huang2017arbitrary}:
\begin{equation}
	AdaIN(x,y) = \sigma(y)(\frac{x-\mu{(x)}}{\sigma(x)})+\mu{(y)}
\end{equation}

where $\mu{(x)}$ and $\mu{(y)}$ denote the mean value of the content and style feature map respectively; $\sigma{(x)}$ and $\sigma{(y)}$ denote the variance value of the content and style feature map respectively.

\begin{table}[h]\tiny
	\centering
	\caption{{Detailed configuration of $\operatorname{StyleAdaIN}$.} }\label{tab:StyleAdaIN}
	\resizebox{\columnwidth}{!}{
		\begin{tabular}{l|cccccc}
			
			\hline									
			Layers	&	$\operatorname{OP}$	&	$\operatorname{IN}$	&	$\operatorname{OUT}$	&	$\operatorname{ACT}$	\\
			\hline									
			\multirow{3}{*}{SplattingBlock1}	&	conv1	&	64	&	64	&	tanh	\\
			&	conv1	&	64	&	256	&	tanh	\\
			&	AdaIN	&	256	&	256	&	-	\\
			\hline									
			\multirow{3}{*}{SplattingBlock2}	&	conv1	&	128	&	128	&	tanh	\\
			&	conv2	&	128	&	256	&	tanh	\\
			&	AdaIN	&	256	&	256	&	-	\\
			\hline									
			\multirow{3}{*}{SplattingBlock3}	&	conv1	&	256	&	256	&	tanh	\\
			&	conv2	&	256	&	256	&	tanh	\\
			&	AdaIN	&	256	&	256	&	-	\\
			\hline									
			\multirow{3}{*}{SplattingBlock4}	&	conv1	&	512	&	512	&	tanh	\\
			&	conv2	&	512	&	256	&	tanh	\\
			&	AdaIN	&	256	&	256	&	-	\\
			\hline									
			
		\end{tabular}
	}
	\vspace{-2mm}
\end{table}

\begin{table}[h]\tiny
	\centering
	\caption{{Detailed configuration of $\operatorname{Clasifier}$.} }\label{tab:Clasifier}
	\resizebox{\columnwidth}{!}{
		\begin{tabular}{l|cccccc}
			
			\hline									
			Layers	&	$\operatorname{OP}$	&&	$\operatorname{IN}$	&	$\operatorname{OUT}$	&	$\operatorname{ACT}$	\\
			\hline									
			Conv1	&	Conv	&&	1024	&	512	&	tanh	\\
			Conv2	&	Conv	&&	512	&	1024	&	tanh	\\
			Conv3	&	Conv	&&	1024	&	512	&	tanh	\\
			
			Conv4	&	Conv	&&	512	&	2048	&	-	\\
			\hline

		\end{tabular}
	}
	\vspace{-2mm}
\end{table}

\section{The visualization of 3DLUT parameters}
\begin{figure}[htbp]
	\vspace{1mm}
	\centering
	\includegraphics[width=1 \linewidth]{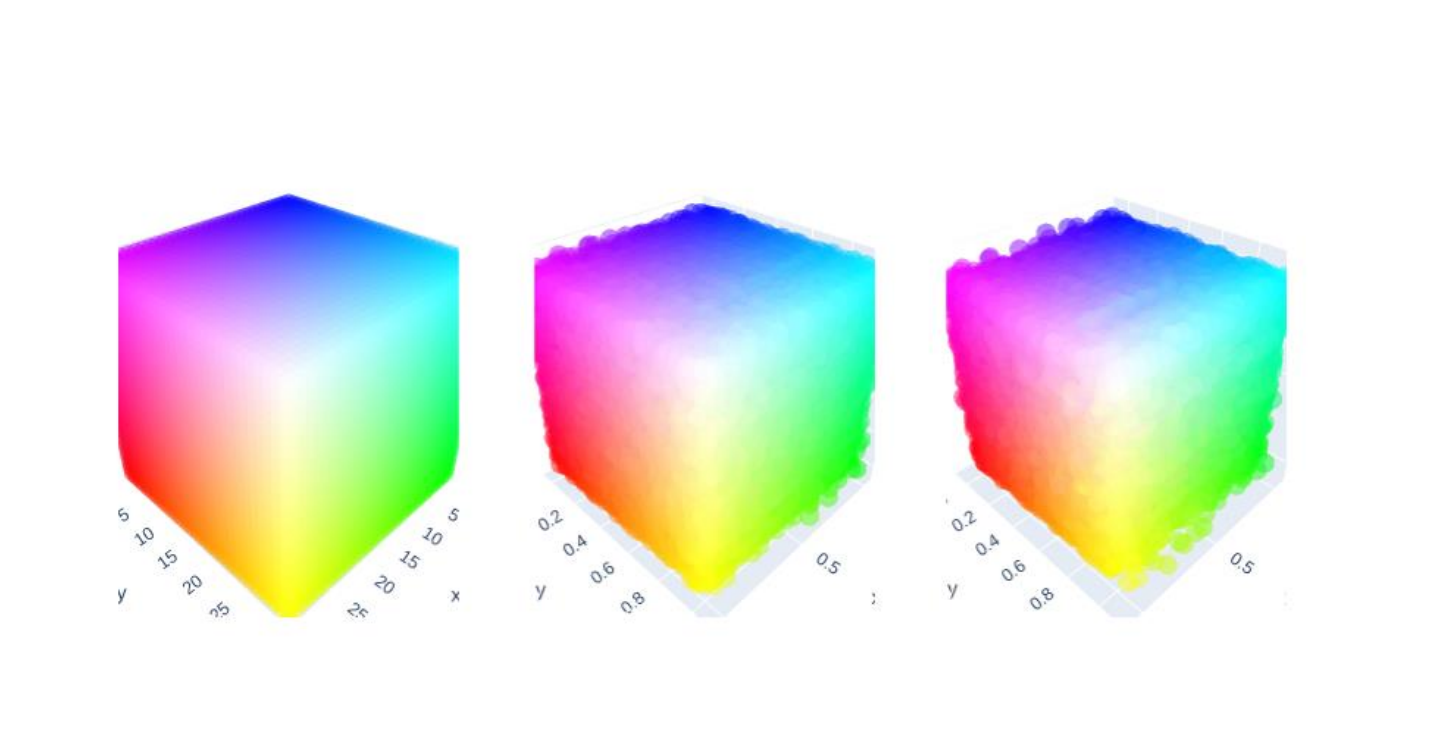}
	\caption{\textbf{The visualization of 3DLUT parameters.} On the left is the 3DLUT in its initial state, in the middle is the 3DLUT trained on a large dataset, and on the right is the 3DLUT trained on a specific style image and a content image.}
	\label{fig:sup_3dLUTs}
	\vspace{-4mm}
\end{figure}

To better understand the effect of 3DLUT parameters pre-training on large-scale datasets and test-time training in specific content image and style image, we show the visualization results of 3DLUT after reconstruction in Fig.~\ref{fig:sup_3dLUTs}. On the left is the 3DLUT in its initial state, in the middle is the 3DLUT trained on a large dataset, and on the right is the 3DLUT trained on a specific style image and a content image. Although it cannot replace the status of all LUTs, at least it can be seen from the figure that the parameters of the 3DLUT have changed after the test-time training.

\section{Additional visual results }

\begin{figure}[htbp]
	\vspace{1mm}
	\centering
	\includegraphics[width=1 \linewidth]{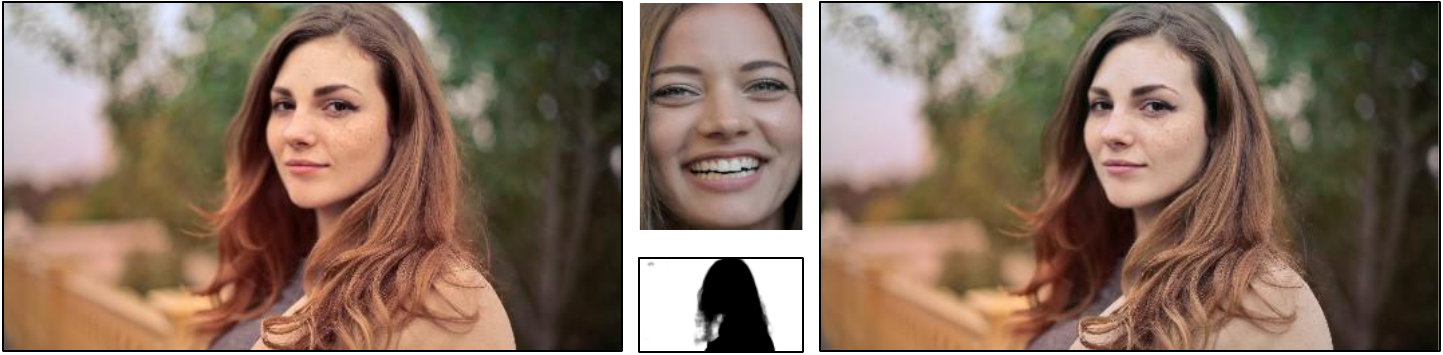}
	\caption{\textbf{Photorealistic stylization results for skin color transfer.} The left is the original content image; the upper part in the middle is the style image, the lower part in the middle is the human transparent mask of the original content image; and the right part is the image after skin color transfer using our method.}
	\label{fig:sup_skin_color_transfer}
	\vspace{-4mm}
\end{figure}

Fig.~\ref{fig:sup_skin_color_transfer} shows the application of our method in the task of skin color transfer. We first use the RobustVideoMatting~\cite{lin2022robust} to pick out the portrait from the content image, then transfer the color from the style image to the human image, and finally paste the transferred portrait back to the background of the content image. From the figure, we can see that the girl's face in the stylized image becomes lighter according to the style image. This proves that our method also has potential applications in skin color transfer.

\begin{figure*}[htbp]
	\centering
	\includegraphics[width=1.\linewidth]{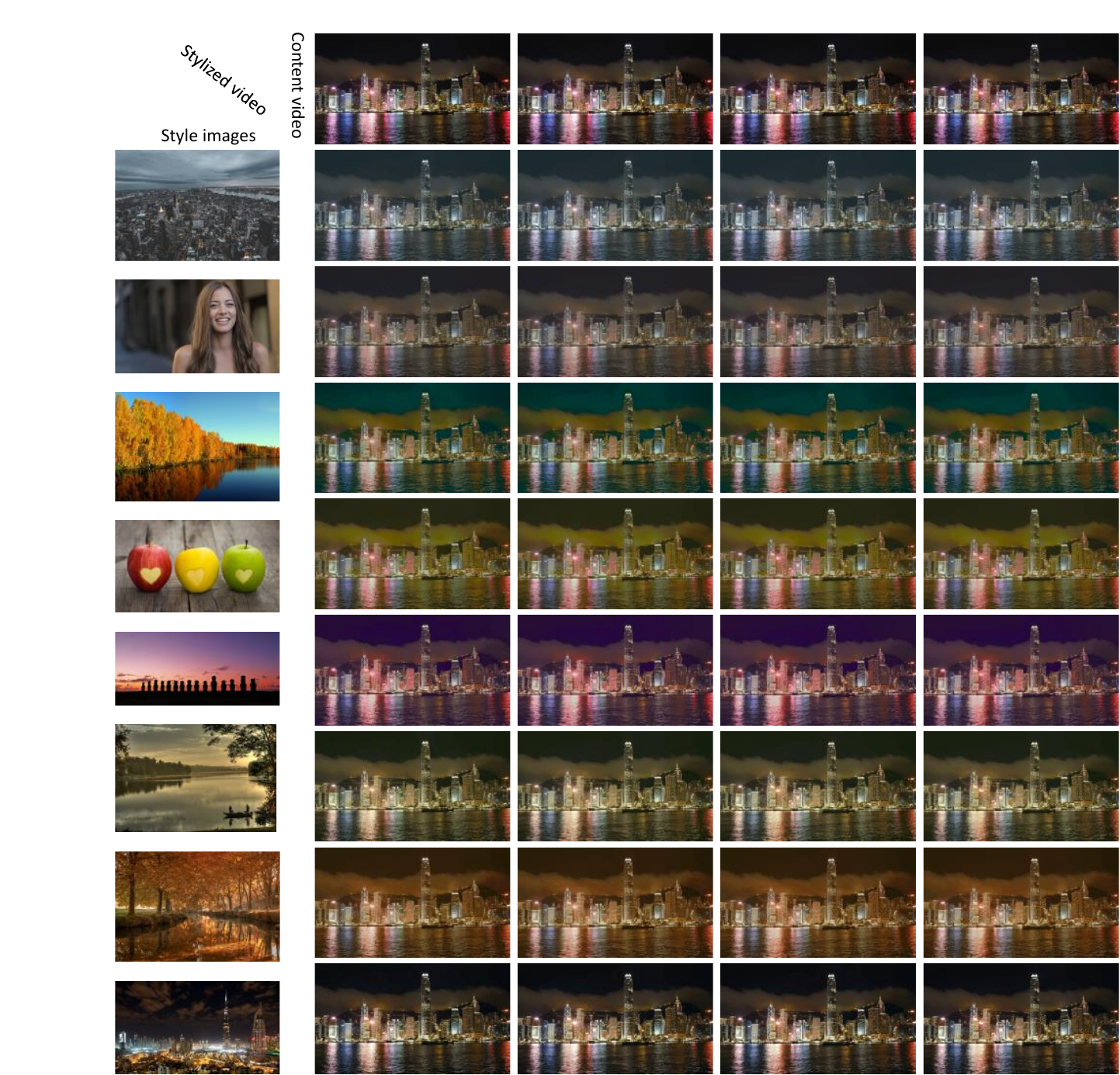}
	\caption{Photorealistic stylization results with different style images on the $\operatorname{night}$ video.}
	\label{fig:sup_night}
	\vspace{-4mm}
\end{figure*}

\begin{figure*}[htbp]
	\centering
	\includegraphics[width=1.\linewidth]{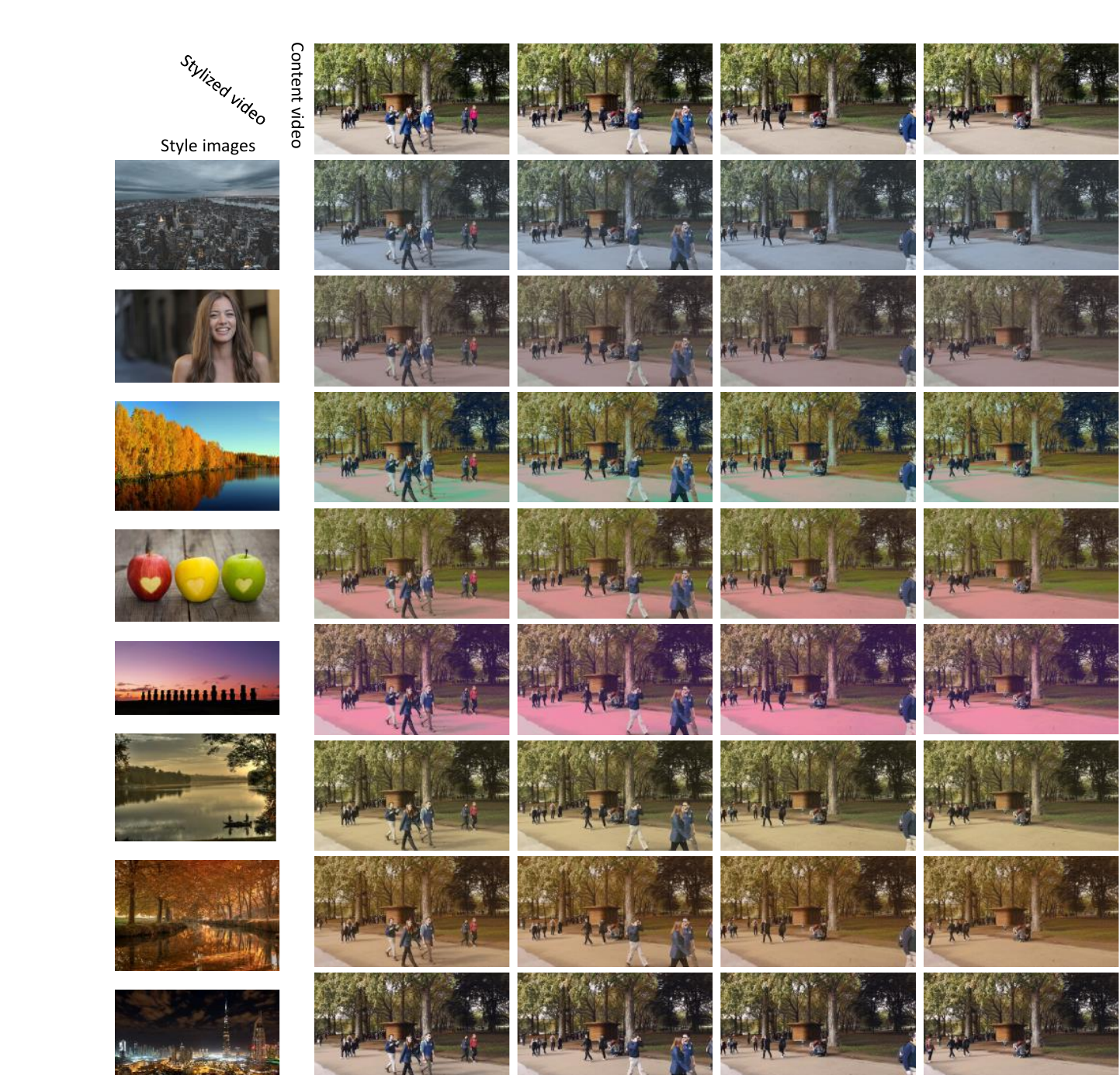}
	\caption{Photorealistic stylization results with different style images on the $\operatorname{pedestrian}$ video.}
	\label{fig:sup_pedestrian}
	\vspace{-4mm}
\end{figure*}

\begin{figure*}[htbp]
	\centering
	\includegraphics[width=1.\linewidth]{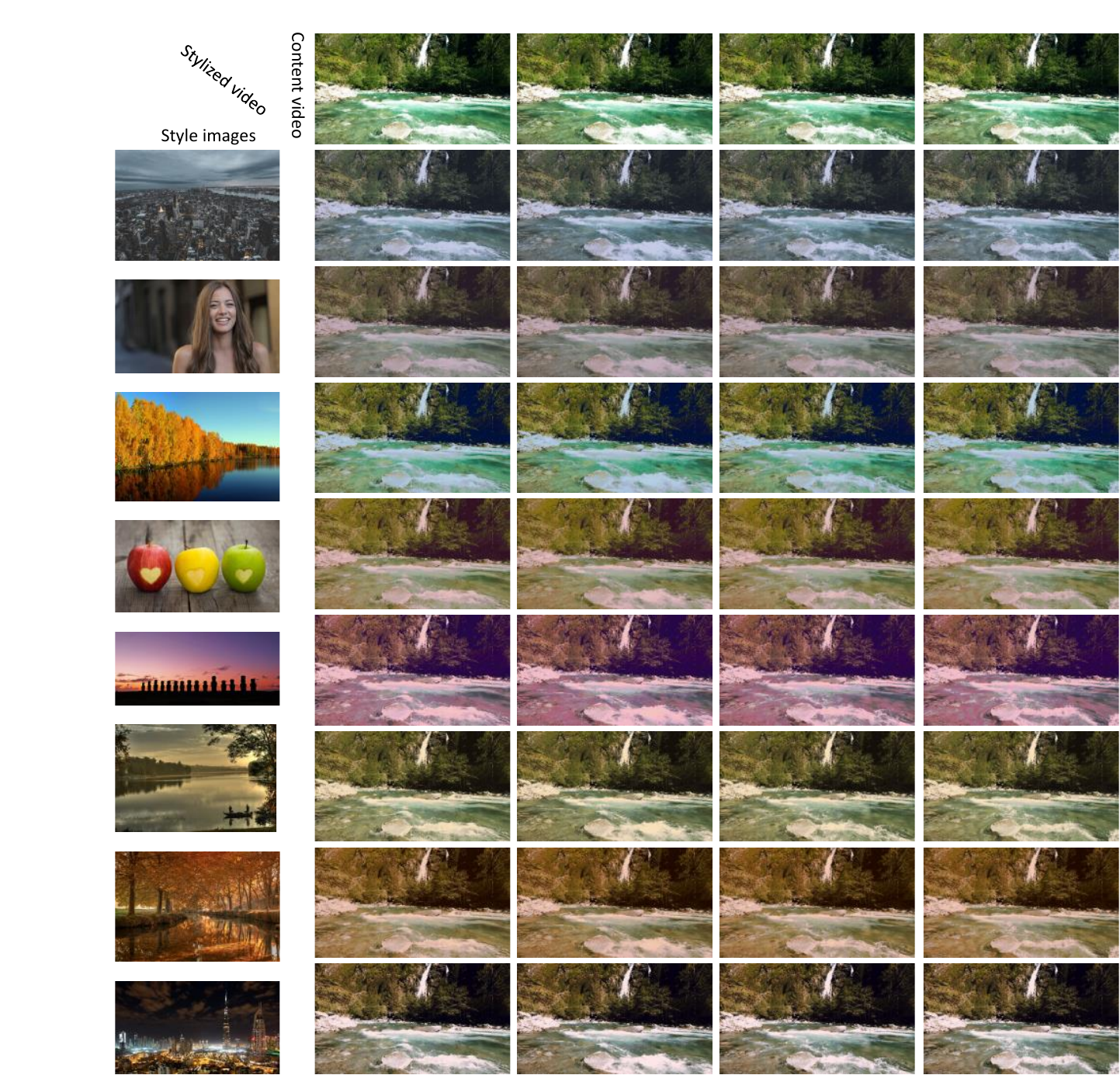}
	\caption{Photorealistic stylization results with different style images on the $\operatorname{stream}$ video.}
	\label{fig:sup_stream}
	\vspace{-4mm}
\end{figure*}

\begin{figure*}[htbp]
	\centering
	\includegraphics[width=1.\linewidth]{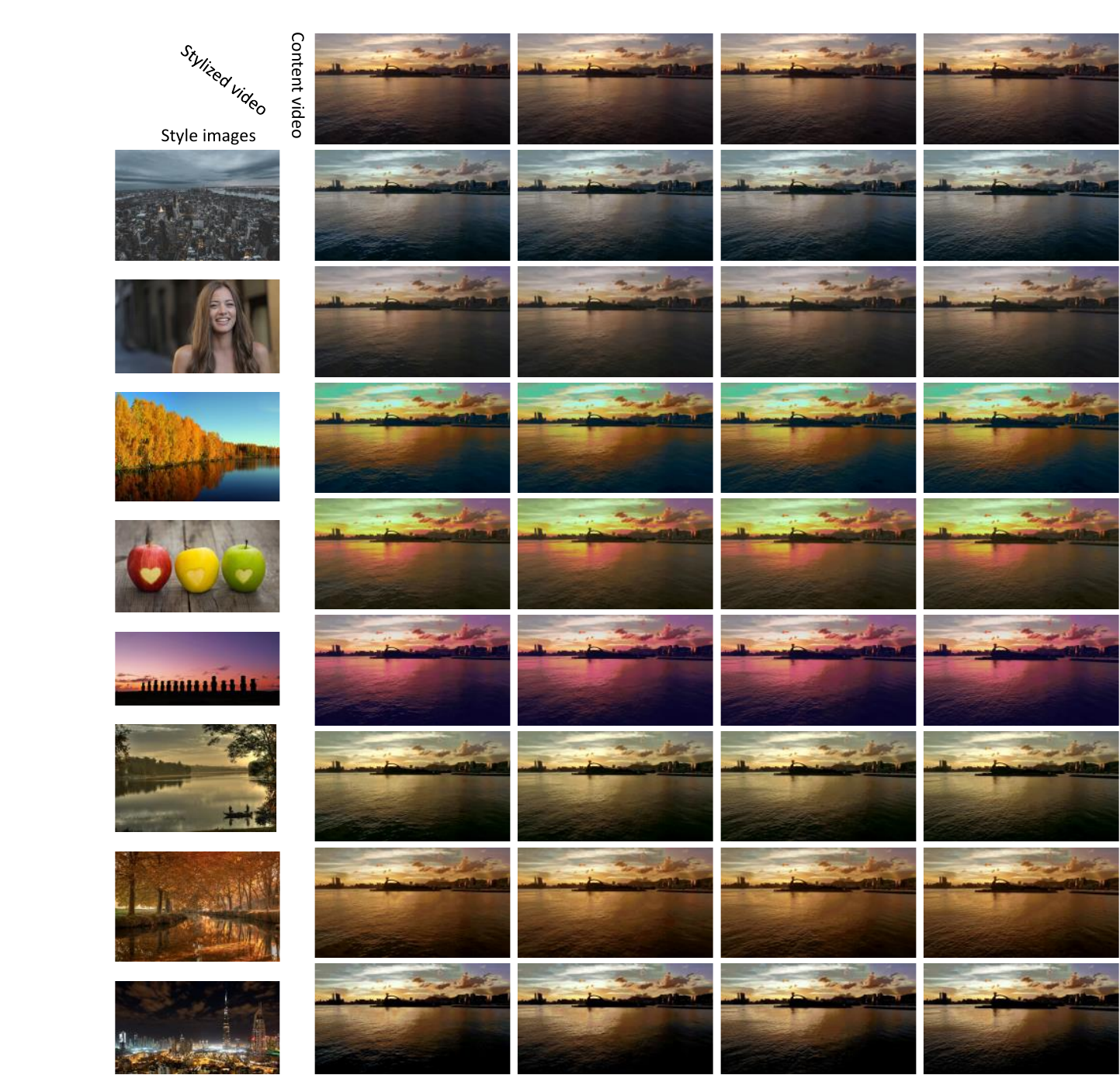}
	\caption{Photorealistic stylization results with different style images on the $\operatorname{sunset}$ video.}
	\label{fig:sup_sunset}
	\vspace{-4mm}
\end{figure*}

Fig.~\ref{fig:sup_night},~\ref{fig:sup_pedestrian},~\ref{fig:sup_stream} and~\ref{fig:sup_sunset} shows more photorealistic stylization results with different style images on the $\operatorname{night}$, $\operatorname{pedestrian}$, $\operatorname{stream}$, and $\operatorname{sunset}$ videos.

\end{document}